\documentclass[11pt,twocolumn,letterpaper]{article}

\usepackage[left=0.8in,right=0.8in,top=0.8in,bottom=0.8in]{geometry}

\usepackage{times}
\usepackage{epsfig}
\usepackage{epstopdf}
\usepackage{graphicx}
\usepackage{amsmath}
\usepackage{amssymb}
\usepackage{multirow}
\usepackage{etoolbox} 

\usepackage{url}
\usepackage{svg}
\usepackage{subcaption}
\usepackage{dblfloatfix}
\usepackage{comment}
\usepackage{booktabs}
\usepackage{makecell}
\usepackage{color,array}

\newcommand{\eg}{\textit{e.g.}~}
\newcommand{\ie}{\textit{i.e.}~}
\newcommand{\etal}{\textit{et al.}~}

\usepackage[colorlinks,urlcolor=blue,linkcolor=blue,citecolor=blue]{hyperref}

\newcommand{\myteaser}
{   
    \includegraphics[width=\linewidth]{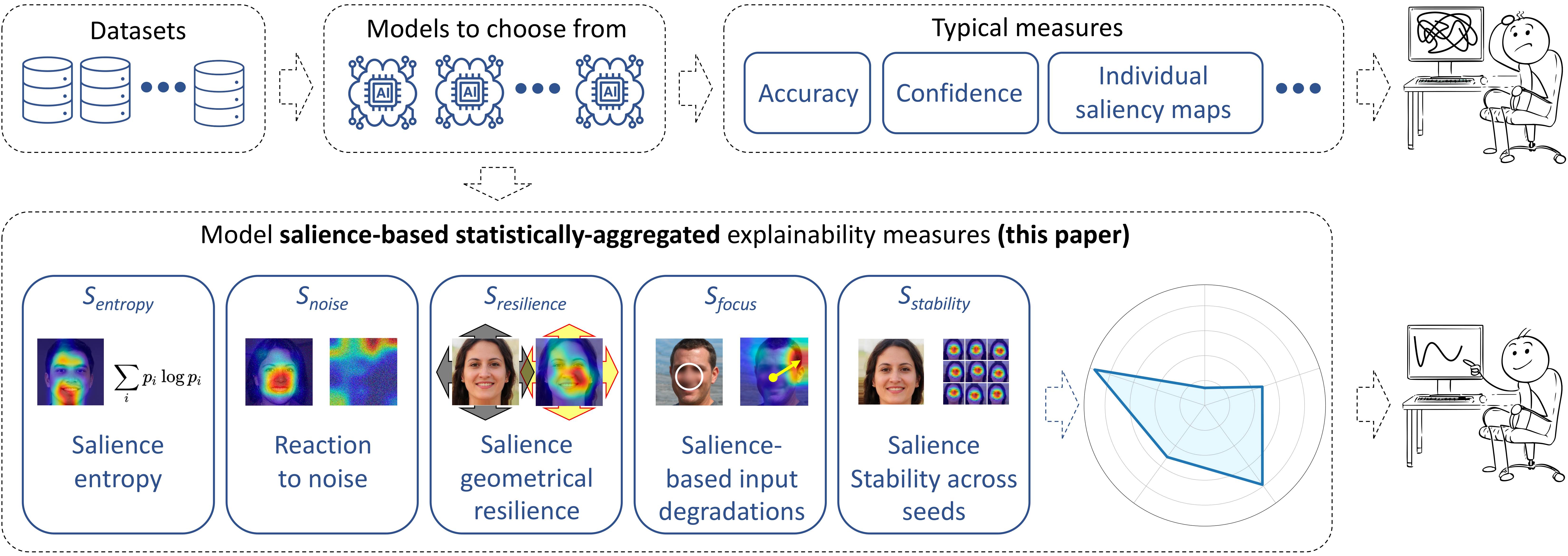}
    Figure 1. An expert-driven model selection, in the case where typical performance metrics are similar for a set of benchmarks, may require multiple views on the same algorithm, aggregated into human-interpretable metrics. This paper leverages the model's salience and -- by aggregating various statistics of saliency maps -- offers a series of metrics explaining the model's behavior. These may serve as exploratory tools to assess the stability of model focus across training runs, data degradation and image spatial transformations, complementing standard performance metrics, such as accuracy or confidence scores.
    \vskip5mm
}

\begin{document}

\makeatletter
\apptocmd\@maketitle{{\myteaser{}\par}}{}{}
\makeatother

\setcounter{figure}{1}


\title{Explain To Me: Salience-Based Explainability\\for Synthetic Face Detection Models} 

\author{Colton~R.~Crum$^{\dag,*}$,~Patrick~Tinsley$^{\dag,*}$,~Aidan~Boyd$^\dag$,~Jacob~Piland$^\dag$,\\Christopher Sweet$^\dag$,~Timothy Kelley$^\ddag$,~Kevin~Bowyer$^\dag$,~Adam~Czajka$^\dag$\\
$^\dag$~University of Notre Dame, Notre Dame, IN 46556, United States\\
{\tt\small \{ccrum,ptinsley,aboyd3,jpiland,csweet1,kwb,aczajka\}@nd.edu}\\
$^\ddag$~Naval Surface Warfare Center -- Crane Division, Crane, IN 47522, United States \\
{\tt\small \{timothy.d.kelley6.civ\}@us.navy.mil}\\
{\footnotesize $^*$ denotes equal contribution}
}


\maketitle


\begin{abstract}


The performance of convolutional neural networks has continued to improve over the last decade. At the same time, as model complexity grows, it becomes increasingly more difficult to explain model decisions. Such explanations may be of critical importance for reliable operation of human-machine pairing setups, or for model selection when the "best'" model among many equally-accurate models must be established. Saliency maps represent one popular way of explaining model decisions by highlighting image regions models deem important when making a prediction. However, examining salience maps at scale is not practical. In this paper, we propose five novel methods of leveraging model salience to explain a model behavior at scale. These methods ask: (a) what is the average entropy for a model's salience maps, (b) how does model salience change when fed out-of-set samples, (c) how closely does model salience follow geometrical transformations, (d) what is the stability of model salience across independent training runs, and (e) how does model salience react to salience-guided image degradations. To assess the proposed measures on a concrete and topical problem, we conducted a series of experiments for the task of synthetic face detection with two types of models: those trained traditionally with cross-entropy loss, and those guided by human salience when training to increase model generalizability. These two types of models are characterized by different, interpretable properties of their salience maps, which allows for the evaluation of the correctness of the proposed measures. We offer source codes for each measure along with this paper.

\end{abstract}


\section{Introduction}


The explanation of AI decisions to humans, while still an emerging area, is becoming a crucial element of neural network training and evaluation.
In domains such as health, law, and finance, the ability to extract semantically meaningful information from a model is oftentimes as important as model performance. An early opinion on the applicability of AI in these domains comes from the UK House of Lords' report \cite{UK_HoL_2018}, which reads: ``We believe it is not acceptable to deploy any artificial intelligence system which could have a substantial impact on an individual’s life, unless it can generate a full and satisfactory explanation for the decisions it will take. In cases such as deep neural networks, where it is not yet possible to generate thorough explanations for the decisions that are made, this may mean delaying their deployment for particular uses until alternative solutions are found.'' This sentiment has spurred the recent quest for methods that offer human-explainable reasoning behind how a model makes its final decision, while still maintaining model performance.

The most straightforward way of judging model explainability is to study \textit{model salience}. In vision-related tasks, model salience is a visualization technique that highlights where a model focuses when making a decision. In principle, this is done by examining  model output sensitivity to local changes made to the input. For black-box models, these changes can be random \cite{Fong_ICCV_2017}, while for white-box models, model gradients change with respect to a given input; this is explained in further detail in Sec. \ref{Model Salience as Explainable AI}). 
These extracted {\it salience maps}, also known as {\it class activation maps}, are useful in explaining which regions of a given image are important in the decision-making process. However, given the sheer volume of image data in vision-related tasks, it is neither practical nor feasible for humans observe every salience map by hand. 
In this work, {\bf the proposed five methods distill meaningful information from salience maps to aid humans in making informed decisions in the model selection process}. This is useful in situations where a set of models perform similarly on a given benchmark (see Fig. 1). More specifically, the proposed methods can justify preference in certain models by measuring:

\begin{itemize} 

\item[(a)] the complexity of the salience maps through their entropy; well-focused models should generate lower-entropy salience maps (Sec. \ref{Salience Entropy});

\item[(b)] the model's {\it reaction to nonsense} by observing salience for random inputs; salience maps should (on average) be less focused when the model is fed nonsense, compared to genuine inputs (Sec. \ref{Salience Randomness});

\item[(c)] how model salience changes after applying selected Euclidean transformations to model input; the salience of well-trained models should (on average) follow the same transformations (Sec. \ref{Salience Augmentations});

\item[(d)] how model salience changes when salient and non-salient regions of an input image are degraded; well-trained models should react intuitively when degradations are applied to salient and non-salient image regions; intuition suggests degradations to \textit{salient} regions should translate to large alterations of focus, while degradations to \textit{non-salient} regions should minimally impact model behavior (Sec. \ref{Salience Degradation});

\item[(e)] how model salience changes across independently trained models when sharing the same model architecture, training strategy, and training data; well-trained models should demonstrate high salience stability across multiple training runs, suggesting that -- independently of random initialization -- the models converge toward the same salient information when solving a given visual task (Sec. \ref{Salience Stability}). 

\end{itemize}

To concretize this work, the proposed measures are applied for the biometric task of synthetic face detection, in which several recent Generative Adversarial Networks (GANs) are used to generate images. To further evaluate the proposed measures, two types of models are tested: (a) those trained traditionally (with cross-entropy loss), and (b) those trained in a recently proposed human salience-guided manner, which increases the model's generalization capabilities, prevents from overfitting, and increases the stability of convergence across training seeds \cite{boyd2022human, Boyd_2023_WACV_CYBORG}. The juxtaposition of the measures for each model type adds to understanding their predictive capabilities.
%
%
For each method, a performance-related metric (Area Under the Curve: AUC) is reported in addition to an SSIM-based explainability-related measure. The latter aggregates salience information into standalone measures, simplifying the judgment about the model usefulness. Source codes facilitating the replicability of this work, and further use of the proposed measures, are offered along with this paper\footnote{\url{https://anonymous.4open.science/r/Explain2Me-0B19/}}.


\section{Related Work}
\subsection{Explainable Biometrics}


Given the prevalence of deep learning-based models in research and practice, the ability to explain a given model's predictive behavior has seen increased emphasis~\cite{ angelov2021explainable, dovsilovic2018explainable, tjoa2020survey}. The common thread among {\it explainable AI} (XAI) research is the goal of understanding how and why models render their ultimate decision. There are currently many post hoc methods to explain models that were trained with model performance in mind rather than human explainability. Two of these methods are the well-known LIME~\cite{lime} and SHAP~\cite{lundberg2017unified} techniques. Goldstein \etal~\cite{goldstein2015peeking} and Molnar \etal~\cite{molnar2020interpretable} also offer partial dependence plots, accumulated local effect (ALE) plots, and individual conditional expectation (ICE) plots towards the common goal of model explainability. Yellowbrick~\cite{bengfort_yellowbrick_2018} and the What-If Toolkit~\cite{wexler2019if} also provide built-in statistical and visualization methods for XAI in scikit-learn and TensorFlow, respectively.

Beyond retrospective post hoc methods, XAI has been enforced proactively in the model building and training process as well as in the documentation. Fact sheets, as seen in ~\cite{arnold2019factsheets, richards2020methodology} serve as an explanatory guide to (i) developers when training models, (ii) dataset curators when releasing public data, and (iii) end users when using the models for their own use. In this way, no models are left unturned; new models can be developed and trained with an explicit emphasis on explainability, and existing models can be assessed and retrofitted to assure explainable behavior. 

\subsection{Model Salience as Explainable AI} \label{Model Salience as Explainable AI}

\paragraph{Salience Map Methods} Model salience has been used extensively in the XAI community to help interpret why models make their decisions. There are many different ways of assessing model salience, such as through neuron activations ~\cite{score_cam, IS_cam, activation_based_cam, SS_cam}, or through the gradients~\cite{layer_cam, axiom_cam, smooth_cam, selvaraju2017grad, chattopadhay2018grad}. One of the most popular salience methods is Gradient-Weighted Class Activation Mappings (Grad-CAMs), which visualizes the gradients of the predicted class at the final convolutional layer. Grad-CAMs are a more sophisticated method of evaluating model salience since they are applicable to CNNs with fully-connected layers and various architecture designs \cite{selvaraju2017grad}. Salience maps are created at the corresponding dimensions of the convolutional feature map (7x7), and are then up-scaled to the original input size (224x224) for ease of viewing.

Further, salience maps can vary in two main regards: (i) to the input data, and (ii) to the map construction method. Naturally, salience maps vary substantially across differing datasets. For example, salience maps generated for frontal face images for face detection will look different when compared to those generated for general object detection. This is due to the fact that model salience is explicitly derived from supplied input images, which differ between datasets. Similarly, human salience also varies in response to the input data. Humans are familiar with identifying both faces and objects in images, but the regions of interest will vary depending on which type of image is presented to a human annotator. 

In the context of constructing salience maps, there are many contributing factors, such as model architecture, how the model is trained (the loss function), which layer of the model the salience is taken from, etc. For human salience, the construction of salience differs dramatically across human annotators, especially in tasks where some are experts while others are not, such as face recognition ~\cite{noyes2017super}. Additionally, human salience can be constructed in different manners: through written annotations (text), drawn annotations, and gaze fixations on an image via eye tracker.


\paragraph{Salience Map Uses} Salience maps are most commonly used to ``hand check'' the model and decide if the model is generalizing appropriately~\cite{selvaraju2017grad, ramaswamy2020ablation, chattopadhay2018grad}. Human examiners can then decide whether or not the salient regions of the model follow conventional feature detection. For example, in the case for real or synthetic face detection, the model is expected to focus on regions related to the face, such as the eyes, nose and mouth. Models that focus on regions similar to humans are more easily understood by humans. If a model has high accuracy but its salience maps focus on irrelevant features, there is a chance that the model has focused on accidental information in the training data and may have trouble generalizing to unseen samples. In this case, the model may be deficient at the task at hand and likely should not be deployed. However, finding irrelevant features is often difficult for humans on certain tasks and can be entirely dataset-dependent.

Recent work has also combined salience maps generated by models and human annotators to improve model performance and explainability.
Boyd \etal features CYBORG, which uses human annotations on faces to penalize the loss function during training~\cite{Boyd_2023_WACV_CYBORG}. Not only did this model outperform standard cross-entropy models, but it even surpassed cross-entropy models trained on seven times the training data. Aswal \etal~\cite{SaliencyMapsWithoutSacrificingAccuracy} used data augmentation techniques and transformations to compare model salience to human annotations while maintaining similar levels of performance accuracy. In \cite{DeepGaze}, Lindardos \etal used model salience to help generalize models and improve overall performance of salience prediction and out-of-domain transfer learning. Human-aided salience maps have also shown to improve the generalizability of neural networks~\cite{boyd2022human}. The large-scale effort to improve salience maps, understand their decisions, and combine them with human intuition motivates the need for a rigorous assessment of these maps.

\section{Proposed Explainability Measures}


Class activation maps (CAMs) or Gradient-Weighted CAMs (Grad-CAMs) are useful when manually checking single images of interest to the model, but fail to capture more general trends in the context a collection of samples across an entire dataset. To remedy this, the measures proposed in this paper more generally assess model salience, relieving operators of the timely task of singular image-by-image observation to understand model behavior at the dataset level. Additionally, the proposed measures allow for any type of salience to be measured and compared so long as they are presented as heatmaps; these include human annotations, eye tracking data, and various types of class activation mappings.

\subsection{Salience Entropy ($S_{entropy}$)}
\label{Salience Entropy}

\begin{figure}[!htb]
  \centering
  \includegraphics[width=\linewidth]{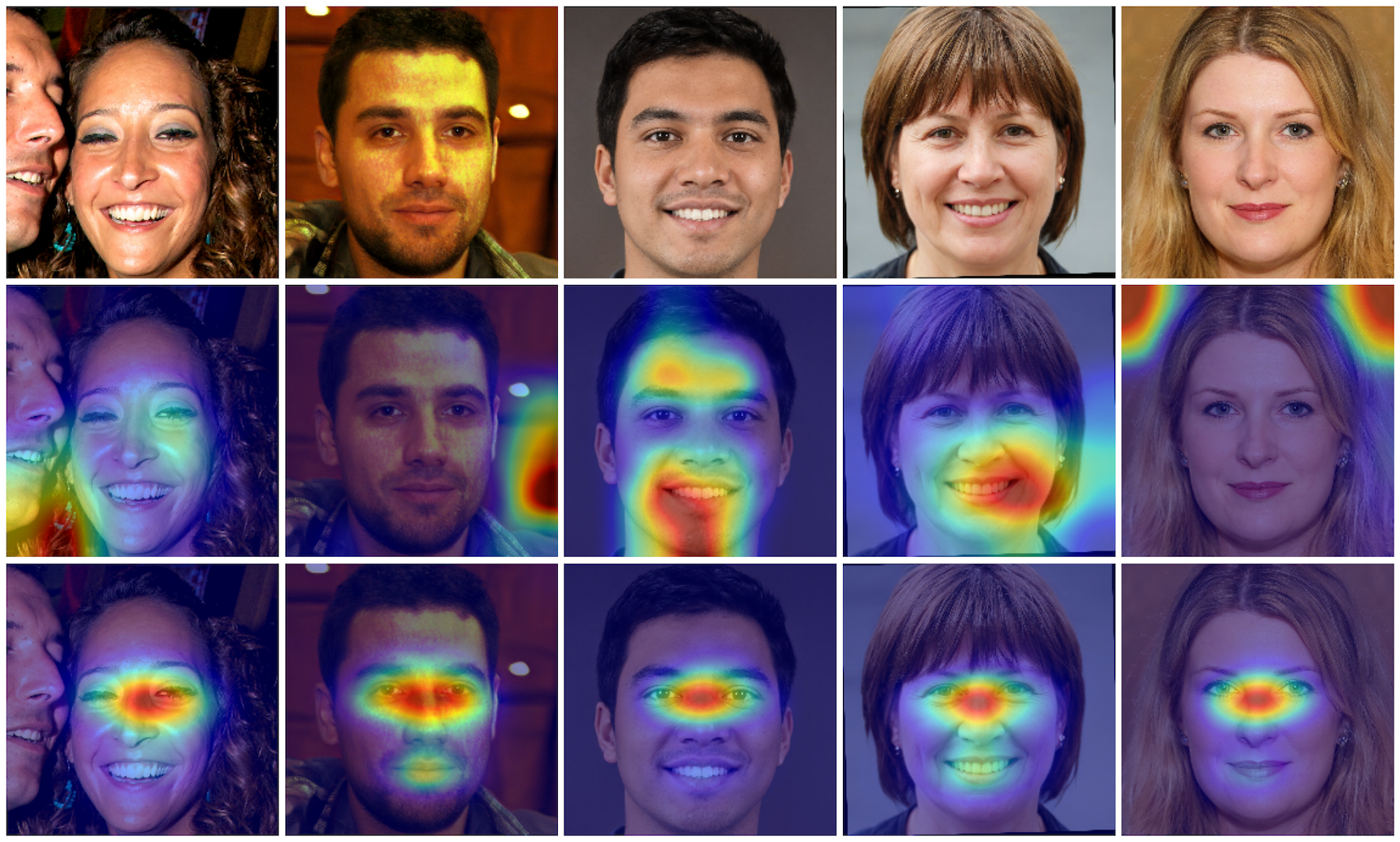}
  \caption{Illustration of $S_{entropy}$ explainability measure based on {\it model's salience entropy}. {\bf Top row, from left to right:} real sample (FFHQ-sourced) and four fake samples (generated by StyleGAN, StyleGAN2, StyleGAN2-ADA, and StyleGAN3). {\bf Middle row:} corresponding salience maps (GradCAMs) for ResNet model trained with cross-entropy loss only. Model focus is indicated by a spectrum color map ranging from blue (low focus) to red (high focus). {\bf Bottom row:} salience maps for ResNet model trained in human-guided manner (CYBORG loss).}
  \label{fig:entropy_illustration}
\end{figure}

\begin{figure*}[t]
    \centering
    \includegraphics[width=0.8\linewidth]{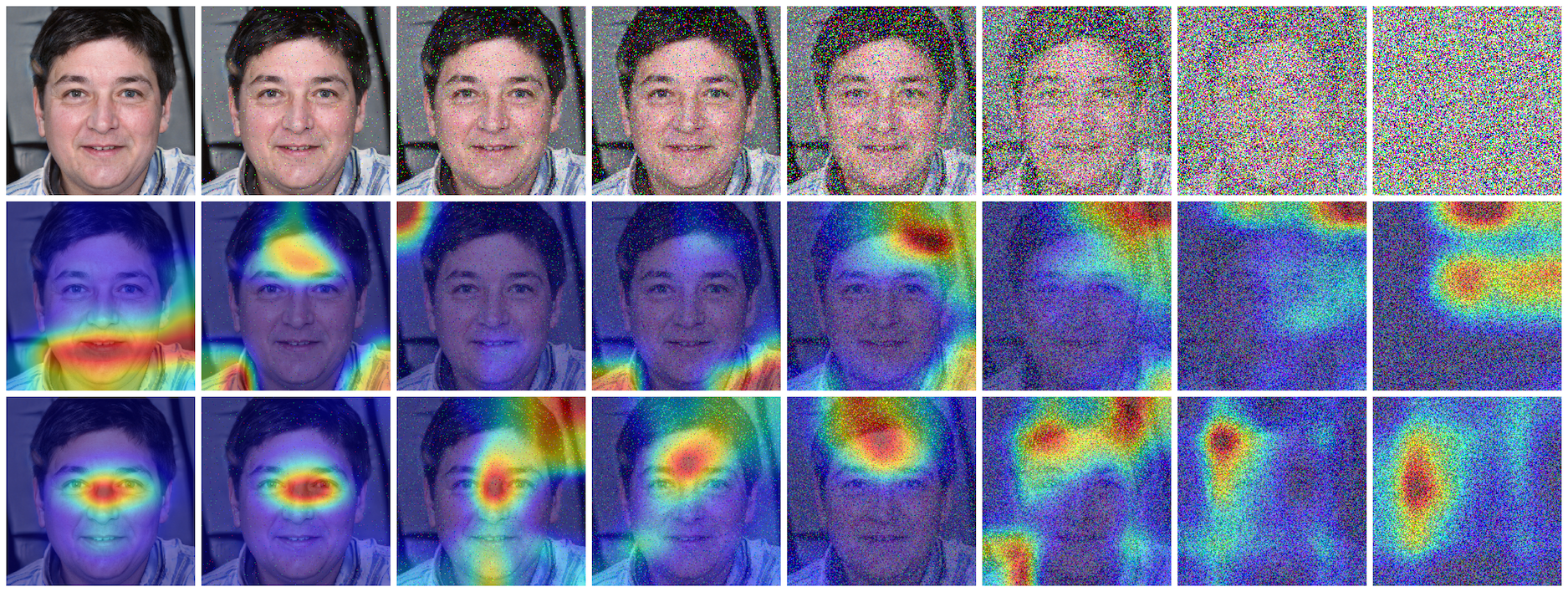}
    \caption{Illustration of the $S_{noise}$ explainability measure assessing model's reaction to noisy inputs. {\bf Top row:} An original input image with increasing intensities of noise, with the leftmost image the original input image with no noise, and the rightmost image an edge case when signal-to-noise ratio equals zero and the model is fed with random uniform noise. Corresponding salience maps for a ResNet model trained traditionally ({\bf middle row}) and with human guidance ({\bf bottom row}) suggest that human-guided training allows for focusing on the actual object for lower signal-to-noise ratios compared to a model trained traditionally. $S_{noise}$ delivers an assessment of this behavior aggregated over an entire dataset.}
    \label{fig:noise_illustration}
\end{figure*}

One of the most straightforward ways of statistically summarizing the model's salience is to study the entropy of generated salience maps. In terms of explainable AI, well-trained models typically have focused salience on the object it is classifying, \eg in synthetic face detection, salience maps should not be randomly-arranged, complex shapes, but should rather be focused around the face region of an image. Fig. \ref{fig:entropy_illustration} illustrates the intuition behind the proposed $S_{entropy}$ measure, in which salience maps are visualized for a few example face images. The model trained in a human-aided way shows more focused salience, which translates to lower $S_{entropy}$ scores, compared to salience maps of the model trained traditionally (without human guidance). High accuracy scores with unfocused salience maps may be indicators that the model has overfit the training data, latching onto extraneous artifacts correlated with class labels during training, but not generalizing well to unseen data.

Formally, we define $S_{entropy}$ as the normalized Shannon's entropy calculated for pixel intensities in the salience map:
\begin{align}
S_{entropy} = -\frac{1}{S_{\mathrm{max}}^{\hat p}(m,n)}\sum_i^I {\bf{p}}(x_i)\log_2 {\bf{p}}(x_i),
\end{align}
\noindent
where ${\bf{p}}(x_i)$ is the estimated probability of a salience map's $i$-th intensity, $I$ is the total number of salience intensities (\eg Grad-CAM histogram bins), and $S_{\mathrm{max}}^{\hat p}(m,n)$ is the maximum entropy for an $n\times m$ image with pixel depth $\hat p$, namely:
\begin{align}
\label{eqn:maxent}\nonumber
S_{\mathrm{max}}^{\hat p}(m,n) &= -\frac{mn}{2^{2\hat p}}\log_2 \left(\frac{mn}{2^{2\hat p}}\right)\times 2^{\hat p} \\
&= -\frac{1}{2^{\hat p}}nm\left(log_2\left( nm\right) - 2\hat p\right),
\end{align}
\noindent
where $\hat p=\min(p,\log_2(nm))$, and $p=8$ for 8-bit salience maps, or $p=\infty$ salience maps represented by real numbers. This normalization factor is a consequence of the observation that the entropy for an image of size $m\times n$ and with pixel depth $p$ is maximized when all of the probabilities ${\bf{p}}(x_i)$ are equal.

The normalization factor $S_{\mathrm{max}}^{\hat p}(m,n)$ plays a key role in getting accurate estimates of the entropy in case of low- and varying-resolution and quantized data (such as Grad-CAMs) and requires a commentary. Rather than binning the salience pixel values to determine the entropy, we consider them to represent an un-normalized probability distribution map \cite{cam_entropy2022}, and the entropy calculated is bound by the size of the input salience map, which differs and depends on the model's last convolutional layer's dimensions (\eg $7\times 7$ for DenseNet and ResNet). The model's salience maps are also often up-sampled to the original input image size (\eg $224 \times 224$) for better visualization purposes. Since this operation does not bring any new information, we calculate the entropy of the maps in their original resolution.

For the common case for arrays of float values, where the pixel depth is infinite, the ``effective'' pixel depth is $\log_2( nm))$ and the maximum entropy Eq. (\ref{eqn:maxent}) can be simplified:
\begin{align}
S_{\mathrm{max}}^{\infty}(m,n) &= \log_2\left(mn\right).
\end{align}
For example, a $7\times 7$ map can only support a pixel depth of $\log_2 7^2$ or 5.614. Therefore, the entropy from each salience map is divided by $S_{\mathrm{max}}^p(m,n) = 5.614$ for this paper's experiments, as DenseNet and ResNet models used in this work have feature map dimensions of $7\times 7$.

\begin{figure}[htb]
    \centering
    \includegraphics[width=\linewidth]{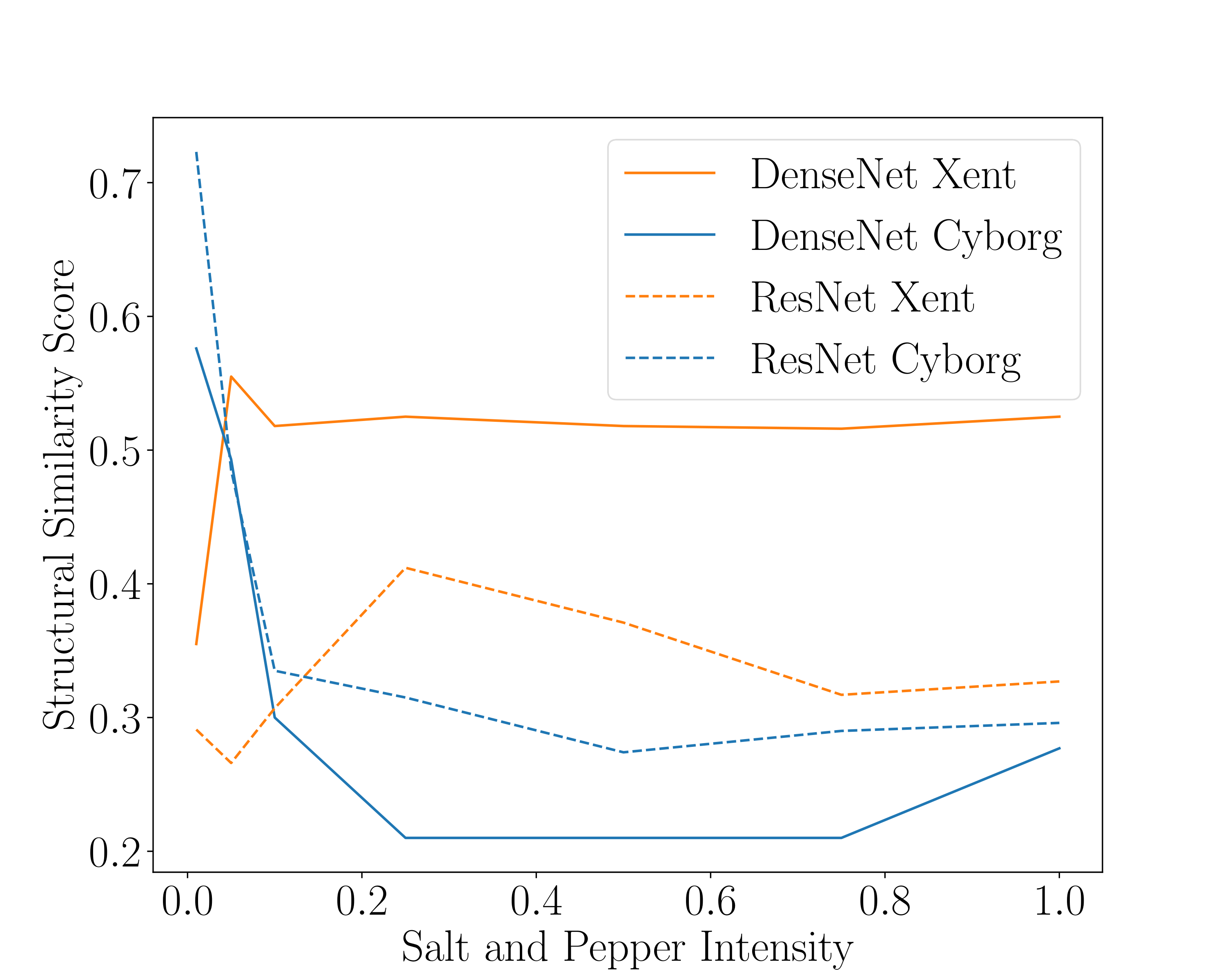}
    \caption{Graphic of the $S_{noise}$ explainability measure comparing the salience map of each model with no noise compared to noisy inputs. Each line represents a model's SSIM score (y-axis) of the original salience map compared to the salience map with noise (x-axis).}
    \label{fig:noise_graphic}
\end{figure}

\begin{figure*}[!htb]
  \centering
  \includegraphics[width=0.9\linewidth]{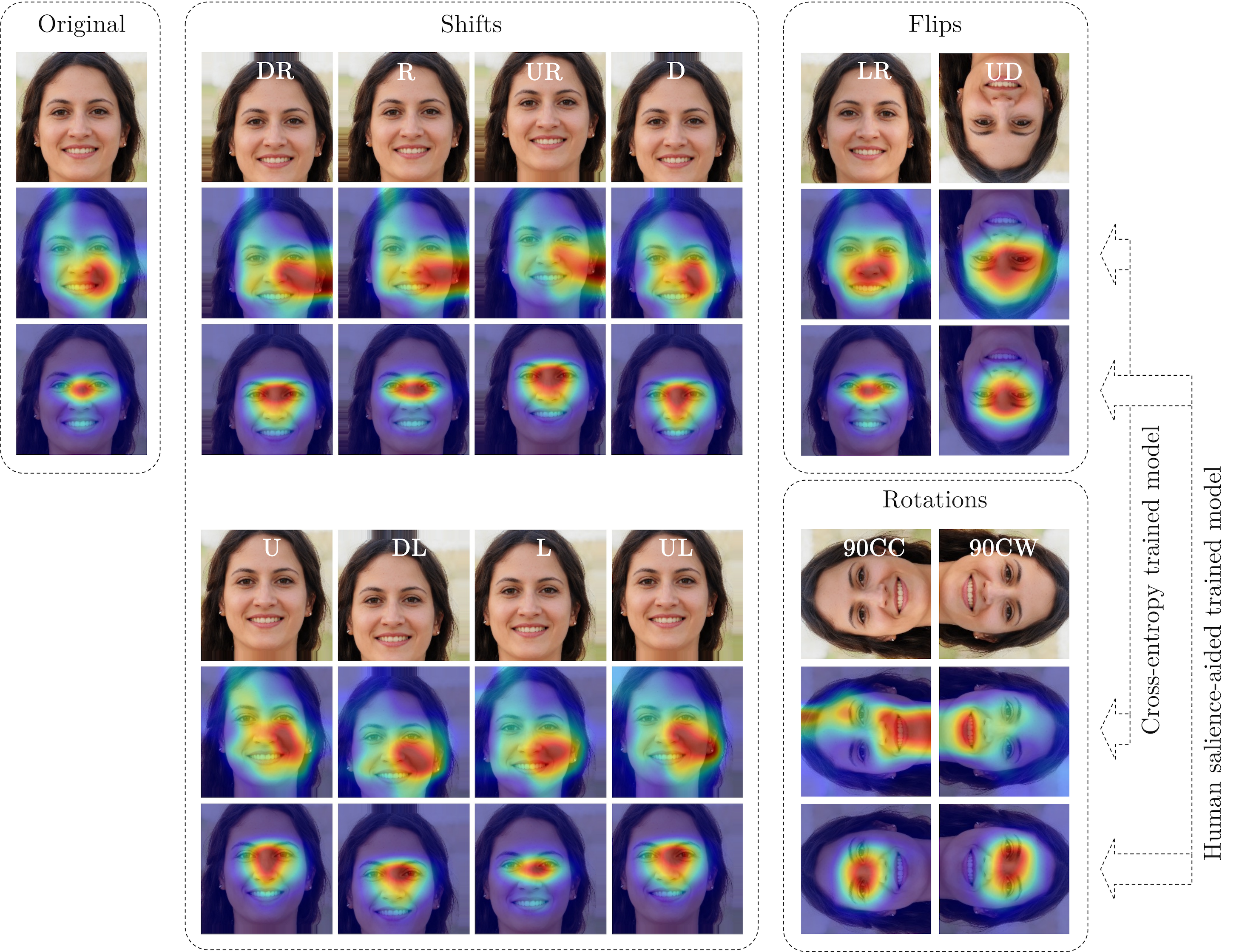}
  \caption{Illustration of the $S_{resilience}$ measuring how the model's salience reacts to selected geometrical transformations: shifts ({DR} = downward-right, {R} = rightward, {UR} = upward-right {D} = downward, {U} = upward, {DL} = downward-left, {L} = leftward, {UL} = upward-left), flips ({LR} = left-over-right, UD = upside-down) and rotations (90CC = 90 degrees counterclockwise, 90CW = 90 degrees clockwise).
  }

  \label{fig:resilience_illustration}
\end{figure*}

\subsection{Salience-Assessed Reaction To Noise ($S_{noise}$)}
\label{Salience Randomness}

It is reasonable to say that well-trained models should degrade their performance gracefully in presence of noise. Typical approaches include recording the model's accuracy or confidence scores as a function of degradation strength. What is proposed in this paper, and may complement the existing measures, is to compare the model's salience maps obtained as the noise is being gradually added to the input, with the salience map calculated for clean samples. Models more robust to a specific noise added should preserve the salience for higher amounts of image degradation. Among various image comparison measures, we propose to adopt a perception-based Structural Similarity Index (SSIM) \cite{Wang_TIP_2004}, initially developed to quantify the perception of errors between distorted and a reference images, and later widely applied as a measure of differences between images:
\begin{equation}
    \mbox{SSIM}(x_1,x_2) = \frac{(2\mu_1\mu_2+\epsilon_1)(2\sigma_{1,2}+\epsilon_2)}{(\mu^2_1+\mu^2_2+\epsilon_1)(\sigma^2_1+\sigma^2_2+\epsilon_2)}
    \label{eqn:ssim}
\end{equation}
\noindent
where $\mu_1$, $\sigma_1$, $\mu_2$, and $\sigma_2$ are mean values and standard deviations of the reference ($x_1$) and distorted ($x_2$) images, respectively, $\sigma_{1,2}$ is a covariance of reference and distorted inputs, and $\epsilon_1$ and $\epsilon_2$ prevent from division by a small denominator. Then:
\begin{equation}
    S_{noise} = \frac{1}{N}\sum_{i=1}^N\mbox{SSIM}(c_i,d_i)
\end{equation}
\noindent
where $N$ is the number of clean-degraded salience map pairs, the reference image $c$ is the salience map obtained for a clean sample, and the distorted image $d$ is the salience map obtained for the degraded image (after adding the noise). Examples of clean and degraded images with their corresponding salience maps are shown in Fig. \ref{fig:noise_illustration}. Additionally, we can measure how quickly the salience map changes with increasing levels of noise. Though this was not one of our core measures of model explainability, we highlight the potential benefits of these curves and the additional information they might bring depending on the domain, as shown in Fig. \ref{fig:noise_graphic}. 

\begin{figure*}
  \centering
  \includegraphics[width=\linewidth]{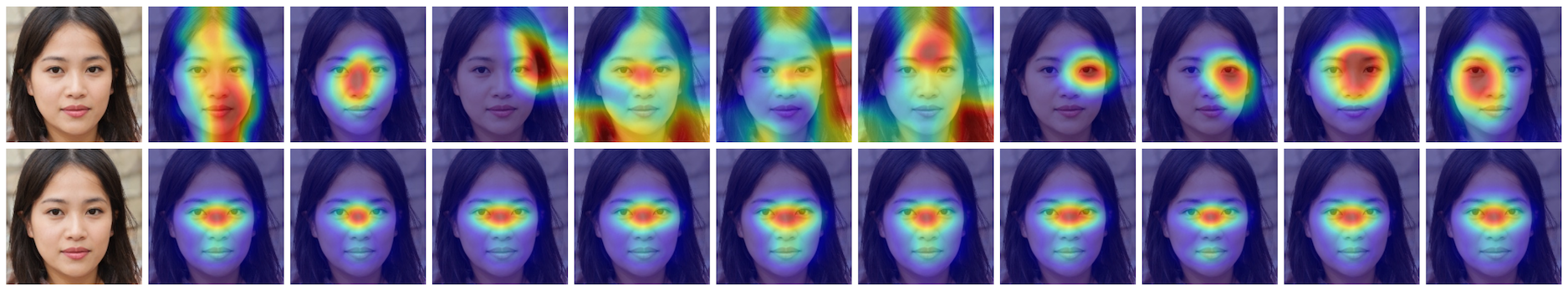}
  \caption{Illustration for the $S_{stability}$ measure assessing the stability of model's salience across training runs. Salience maps for 10 ResNet models trained with cross-entropy loss ({\bf top row}) and with human guidance ({\bf bottom row}) for a given input image shown in the left column.}
  \label{fig:stability_illustration}
\end{figure*}
\begin{figure}
    \centering
    \includegraphics[width=0.6\linewidth]{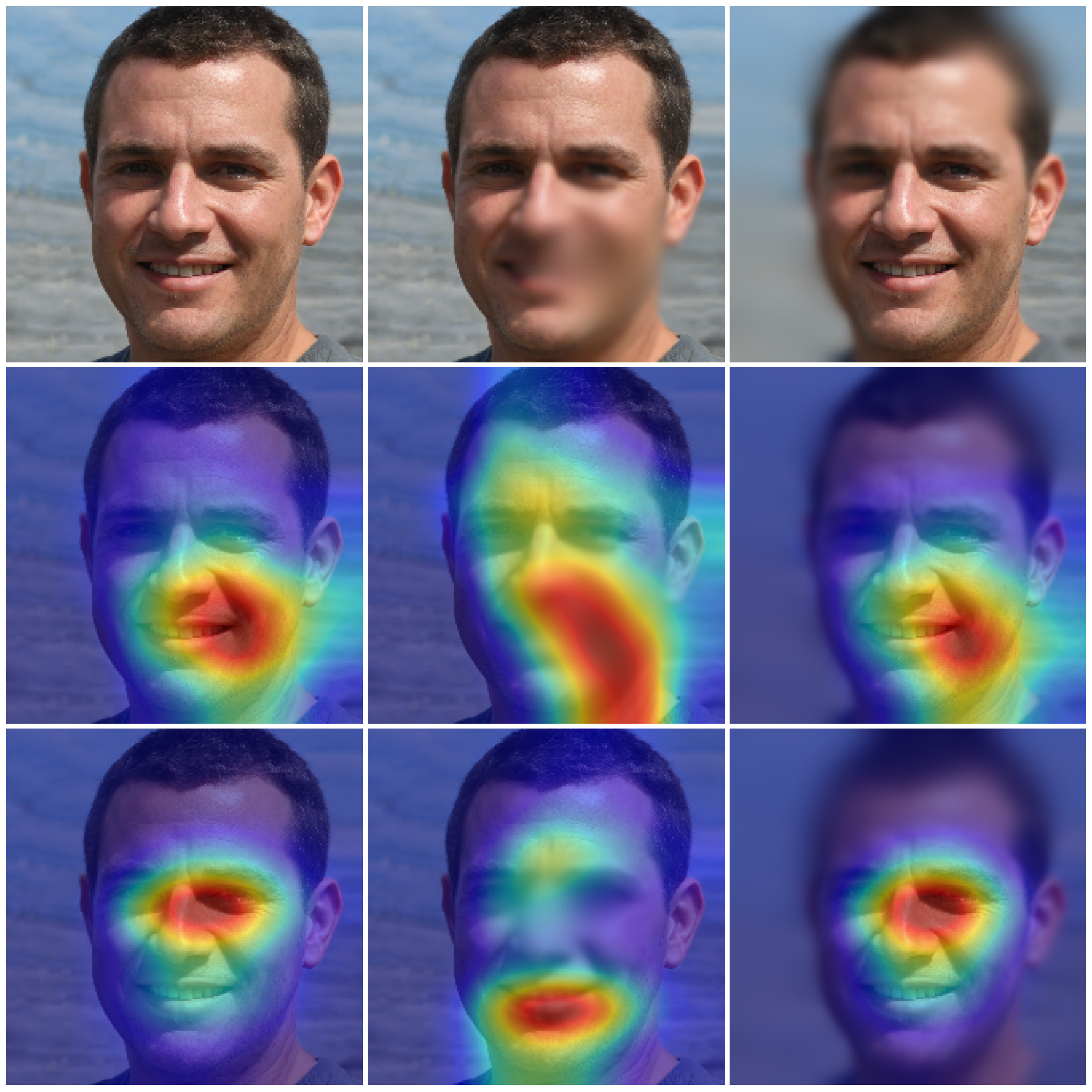}
    \caption{Illustration of the $S_{focus}$ measuring how the model reacts to salience-sourced image degradation. {\bf Top row:} Original image (left) with salient (middle) and non-salient (right) regions ``removed'' by blurring. Corresponding model's salience maps for a ResNet model trained traditionally ({\bf middle row}) and with human guidance ({\bf bottom row}).}
    \label{fig:degradation_illustration}
\end{figure}

\subsection{Salience Resilience to Geometrical Transformations ($S_{resilience}$)}
\label{Salience Augmentations}

The third way we propose to study the model's behavior is to compare the salience maps obtained for a given input image with those obtained for the same image after applying selected geometrical image transformations. The intuition behind this measure is that a model having a good ``understanding'' of the actual (rather than accidentally correlated with class labels) features will follow their geometrical placement in the input data. For this paper, we have three types of transformations: x-y translations, vertical and horizontal flips, and 90-degree rotations, as shown in Fig. \ref{fig:resilience_illustration}. But this measure is certainly not tied to any specific set of transformations, which should be selected appropriately for the task being solved by the model. For instance, in the case of synthetic face detection, trained with 
centrally cropped, upright face images, inputs that deviate significantly from the training data (such as homography or nonlinear transformations) could ``break'' the model. 

Formally, we define $S_{resilience}$ for a transformation $T$ as
\begin{equation}
    S_{resilience,T} = \frac{1}{N}\sum_{i=1}^N\mbox{SSIM}(o_i,f_T(t_i))
\end{equation}
\noindent
where SSIM is defined by Eq. (\ref{eqn:ssim}), $N$ is the number of salience map pairs, the reference image $o_i$ is the salience map obtained for an original sample, the distorted image $t_i$ is the salience map obtained for the transformed image, and $f_T(\cdot)$ denotes the inverse $T$ transformation. More specifically, after each transformation is applied (upper rows in Fig. \ref{fig:resilience_illustration}), a secondary salience map $t_i$ is generated (bottom rows of Fig. \ref{fig:resilience_illustration}). The secondary salience maps $t_i$ are ``corrected'' by reversing the geometrical transformation applied to original image via $f_T(\cdot)$, which are finally compared with maps $o_i$ for original samples using the SSIM measure.

\subsection{Salience-Based Image Degradation ($S_{focus}$)}
\label{Salience Degradation}

The fourth manner in which we measure model's salience-based properties is by comparing salience maps of original input images against images that are degraded based on model salience. In order to produce these latter images, we first generate a salience map for an original image, highlighting where on the image the model focuses during classification. We then use the salience map to blur salient and non-salient regions of the image, producing two new images, one of which has \textit{important} regions blurred out (\textit{salience removal}) and the other has \textit{unimportant} information blurred out (\textit{non-salience removal}). An important note is that we chose to blur regions of salience to remove high frequency information from the image, which allows to not introduce artificial features to the input images. This may happen if we remove information by blackening portions of the image, hence introducing strong image gradients, originally not existing in input samples.

If salient regions of the model are important in the model's decision making process, then removing them from an input image should negatively affect the model's ability to classify the images. Conversely, the removal of non-salient regions of an input image should not significantly affect the model's performance. Accordingly, this metric (in an almost self-referential way) indicates how important model salience is to the model itself. We can then evaluate the model by both observing the selected performance metric (\eg AUROC: Area Under the ROC curve) and SSIM, as defined by Eq. (\ref{eqn:ssim}), post-degradation to the baseline original images.

\subsection{Salience Stability Across Training Runs ($S_{stability}$)}
\label{Salience Stability}

The fifth way to assess model's behaviour is to compare salience maps of independently trained models with the same backbone-training strategy configuration, \ie models that have the same model architecture, loss function, training data, but the training is started with a different seed. This can be helpful to assess, quantitatively, whether a combination of training strategy (including data selection) and model architecture converges to the same salient features. Fig. \ref{fig:stability_illustration} illustrates this concept for $N=10$ independent training runs of the ResNet model, with two different training strategies, as in all previous measures (cross-entropy and human guided). It is clear from this example that human-guided model converges to similar salient features compared to the classically-trained model. Formally, we define $S_{stability}$ as
\begin{equation}
    S_{stability} =\frac{1}{K}\sum_{k=1}^K\frac{2}{N(N-1)}\sum_{i>j}\mbox{SSIM}(x_{k,i},x_{k,j})
\end{equation}
\noindent
where SSIM is defined by Eq. (\ref{eqn:ssim}), $x_{k,i}$ is the salience map corresponding to $k$-th test sample and $i$-th training run, $i,j = 1,\dots,N$, where $N$ is the number of training runs, and $k=1,\dots,K$, where $K$ is the number of test samples. Values closer to 1.0 denote high $S_{stability}$.

\begin{table*}[htb]
\centering
\caption{Values of four measures averaged across all salience maps generated for the four model architecture-training strategy configurations (rows) and for each dataset (column). Plus/minus one standard deviation of the result is given. StyleGAN is shortened to S-GAN in the table for all synthetic datasets. $S_{entropy}$ approximates the complexity of the salience maps, with lower values indicating more focused salience. $S_{noise}$ offers an aggregated assessment of a difference between clean samples (authentic and synthetic faces) and degraded samples (after adding {\bf salt-and-pepper noise}, as shown in middle column in Fig. \ref{fig:noise_illustration}). $S_{focus}$ is shown separately for both variants when the salient regions are removed (blurred) and non-salient regions (background) are removed. $S_{focus}$ values for {\bf non-salient} removals should be {\bf higher}, as information not necessary to complete the task (synthethic face detection) should not be salient for the model. Conversely, $S_{focus}$ values for {\bf salient} removals should be {\bf lower}, as these salient regions should completely disrupt the models predictive ability.}
\label{tab:results}
\begingroup
\setlength{\tabcolsep}{4pt} 
\renewcommand{\arraystretch}{1} 
\small
\begin{tabular}{|c|c|c|c|c|c|}
\hline
 {\bf Measure} &{\bf Training} & {\bf Model} & {\bf Authentic samples} & {\bf Synthetic samples} & {\bf Total}\\
 &{\bf Type} & & {FFHQ} & S-GAN / S-GAN2 / S-GAN2-ADA / S-GAN3 & {\bf Average} \\ 
 \hline\hline

\multirow{4}{*}{$S_{entropy}$} & Cross- & DenseNet   & 0.885 $\pm$ 0.123 & 0.896 $\pm$ 0.084 & 0.894 $\pm$ 0.093\\\cline{3-6}
 & entropy & ResNet & 0.884 $\pm$ 0.095 & 0.873 $\pm$ 0.074 & 0.875 $\pm$ 0.079\\
  \cline{2-6}
 & Human & DenseNet & {0.837$\pm$ 0.119} & {0.808 $\pm$ 0.108} & 0.813 $\pm$ 0.111 \\\cline{3-6}
 & salience & ResNet & 0.887 $\pm$ 0.065 & 0.852 $\pm$ 0.066 & 0.859 $\pm$ 0.068 \\
 \hline\hline

 \multirow{4}{*}{$S_{noise}$} & Cross- & DenseNet & 0.493$\pm$ 0.147 & 0.570 $\pm$0.131 & 0.555$\pm$0.139 \\\cline{3-6}
 & entropy & ResNet & 0.322$\pm$ 0.150 & 0.252 $\pm$0.133 & 0.265$\pm$0.141 \\
  \cline{2-6}
 & Human & DenseNet & 0.457$\pm$ 0.146 & 0.503 $\pm$0.131 & 0.494 $\pm$0.135 \\\cline{3-6}
 & salience & ResNet & 0.456$\pm$ 0.159 & 0.492 $\pm$ 0.139 & 0.485$\pm$ 0.145 \\
 \hline\hline

 &\multicolumn{5}{l|}{{\bf Salience removal}}\\ \cline{2-6}
 \multirow{10}{*}{$S_{focus}$}& Cross- & DenseNet  & 0.570$\pm$0.258 & 0.484 $\pm$ 0.278 & 0.501$\pm$0.278 \\ \cline{3-6}
 & entropy & ResNet & 0.692$\pm$0.185 & 0.700 $\pm$ 0.171 & 0.698$\pm$0.174 \\
  \cline{2-6}
 & Human & DenseNet & 0.534$\pm$0.166 & 0.605$\pm$0.162 & 0.591$\pm$0.165 \\\cline{3-6}
 & salience & ResNet & 0.679$\pm$0.140 & 0.776 $\pm$ 0.106 & 0.757$\pm$0.121 \\
\cline{2-6}
 &\multicolumn{5}{l|}{{\bf Non-salience removal}}\\ \cline{2-6}
 & Cross- & DenseNet & 0.817$\pm$0.137 & 0.857$\pm$0.087 & 0.849$\pm$0.101 \\\cline{3-6}
 & entropy & ResNet & 0.706$\pm$0.142 & 0.703$\pm$0.135 & 0.703$\pm$0.137 \\
\cline{2-6}
 & Human & DenseNet & 0.625$\pm$0.190 & 0.703$\pm$0.145 & 0.687$\pm$0.159 \\\cline{3-6}
 & salience & ResNet & 0.758$\pm$0.121 & 0.786$\pm$0.102 & 0.781$\pm$0.107 \\
 \hline\hline

  \multirow{4}{*}{$S_{stability}$}&Cross- & DenseNet & 0.696 $\pm$ 0.191 & 0.760 $\pm$ 0.127 & 0.747 $\pm$ 0.144 \\\cline{3-6}
 & entropy & ResNet & 0.334 $\pm$ 0.152 & 0.406 $\pm$ 0.135 & 0.392 $\pm$ 0.142\\
  \cline{2-6}
 & Human & DenseNet & 0.884 $\pm$ 0.115 & 0.931 $\pm$ 0.061 & 0.922 $\pm$ 0.077 \\\cline{3-6}
 & salience & ResNet & 0.879 $\pm$ 0.129 & 0.941 $\pm$ 0.073 & 0.929 $\pm$ 0.091 \\
 \hline
\end{tabular}
\endgroup
\end{table*}


\section{Evaluation Environment}
\subsection{Datasets} 
\label{Datasets}

\begin{figure}[!t]
    \centering
    \includegraphics[width=\linewidth]{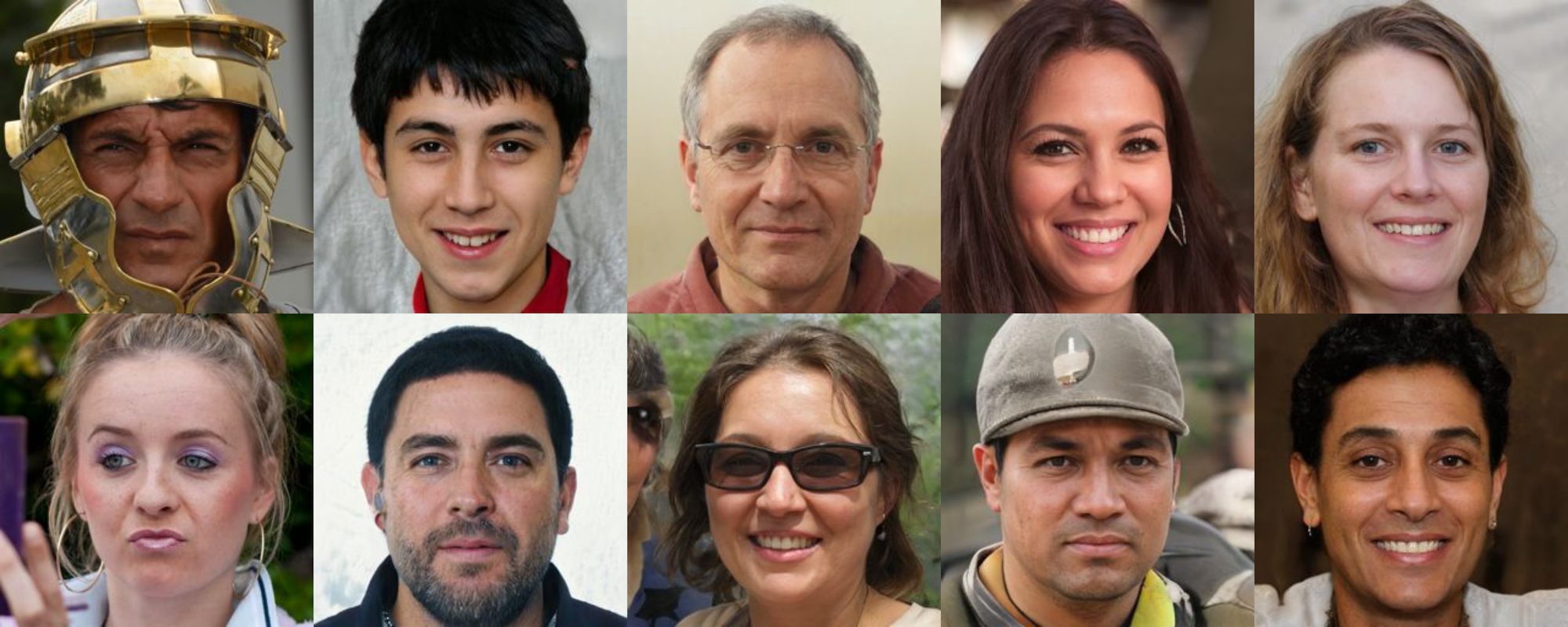}
    \caption{Two examples of face images from each dataset. Columns from left to right: FFHQ, StyleGAN, StyleGAN2, StyleGAN2-ADA, and StyleGAN3.}
    \label{fig:data}
\end{figure}

Two types of datasets were used to demonstrate the efficacy of our proposed explainability metrics: authentic (real) face images pulled from the FFHQ dataset, and synthetic (fake) face images produced by the StyleGAN family of generators. Each generative model was pre-trained on the FFHQ dataset and supplied as part of the official repositories.

\paragraph{Flickr-Faces-HQ (FFHQ)}~\cite{karras2019style} consists of 70,000 high resolution face images collected by NVIDIA via web crawler. The images vary in race, age, background scene, face pose, and fashion accessories. In this work, we randomly selected 10,000 images from the full dataset to create our subset.

\paragraph{StyleGAN Generators} currently consist of four available model types: StyleGAN~\cite{karras2019style}, StyleGAN2~\cite{StyleGAN2}, StyleGAN2-ADA~\cite{Karras2020ada}, and StyleGAN3~\cite{StyleGAN3}. With each official repository, the authors at NVIDIA supply a generator model pre-trained on the FFHQ face dataset. For each model, we produced 10,000 face images with random seeds. In terms of face quality, StyleGAN2, StyleGAN2-ADA, and StyleGAN3 versions supersede the original StyleGAN; StyleGAN was susceptible to ``water bubble'' and edge-based artifacts. 
However, across all iterations of the StyleGAN technology, there exist abnormal artifacts of the synthesis process. These may include disappearing shoulders and misshapen ears (StyleGAN1 - top, bottom), glasses embedded in facial tissue and melted background people (StyleGAN2 - top, bottom), structurally impossible jewelry and unseen head wear accessories (StyleGAN2-ADA - top, bottom), and off-center teeth (StyleGAN3 - top, bottom). Examples of these synthetic artifacts can be seen in Fig.\ref{fig:data}.

\subsection{Model Architectures}

We evaluate our proposed explainability metrics using two popular backbones for deep neural networks: DenseNet121~\cite{huang2017densely} and ResNet50~\cite{ResNet}. These two backbones were chosen primarily due to the fact that they are both available with cross-entropy loss and CYBORG loss (human salience-guided).

\subsection{Training Strategies} 

For each of the two network architectures (DenseNet and ResNet), models are trained using two different loss functions during training: cross-entropy loss and CYBORG loss~\cite{Boyd_2023_WACV_CYBORG}. As proposed by Boyd \etal, CYBORG improves model generalizability by incorporating human-annotated salience into the model's loss function in order to guide the model towards human-judged regions of importance. This is done by juxtaposing the model's salience (Class Activation Mappings) and collected human annotations. The training strategy is particularly relevant in the evaluation of our metrics since we can directly compare how model salience changes across the two loss functions: human-guided (CYBORG or ``CYB'') loss and traditional (cross-entropy or ``Xent'') loss. It is important to note that the models used in this paper are not directly comparable to the original CYBORG paper because only a subset of the test set is used.

\subsection{Model Salience Estimation} 

For our experiments, we use one of the popular methods of evaluating model salience: Gradient-Weighted Class Activation Mappings (Grad-CAMs)~\cite{selvaraju2017grad}, as discussed previously in Section 2.2. 

\section{Evaluation Results}
\label{sec:results}

\begin{table*}
\centering
\caption{Values of $S_{resilience}$ averaged across all salience maps generated for the four model architecture-training strategy configurations (rows) and for each dataset (column), and averaged across genuine and synthetic face images. Meaning of geometrical transformations: {\bf shifts} ({DR} = downward-right, {R} = rightward, {UR} = upward-right {D} = downward, {U} = upward, {DL} = downward-left, {L} = leftward, {UL} = upward-left), {\bf flips} ({LR} = left-over-right, UD = upside-down) and {\bf rotations} (90CC = 90 degrees counterclockwise, 90CW = 90 degrees clockwise). All transformations are illustrated in Fig. \ref{fig:resilience_illustration}).}
\label{tab:resilience}

\begingroup
\setlength{\tabcolsep}{4pt} 
\renewcommand{\arraystretch}{1} 
\small
\begin{tabular}{|c|c|c|c|c|}
\hline
 {\bf Training} & {\bf Model} & {\bf Shifts} & {\bf Flips} & {\bf 90-degree Rotations}\\
 {\bf Type} & & (DR+R+UR+D+U+DL+L+UL)/8 & (LR+TB)/2 & (90CW+90CC)/2\\ 
 \hline\hline
 Cross- & DenseNet & 0.825 & 0.690 & 0.852 \\\cline{2-5}
 entropy & ResNet & 0.609 & 0.491 & 0.814 \\
  \hline
 Human & DenseNet & 0.841 & 0.640 & 0.839 \\\cline{2-5}
 salience & ResNet & 0.716 & 0.600 & 0.851 \\
 \hline
\end{tabular}
\endgroup
\end{table*}

To evaluate the usefulness of the proposed measures, we calculated their values for GradCAM-based salience maps for authentic and synthetic samples, and two types of training: classical cross-entropy-based, and human-aided. Two model architectures (DenseNet and ResNet) were also used. In the case of real images, 4,000 images were sub-sampled from our selected 10,000 FFHQ images. In the case of synthetic images, 1,000 images were similarly sampled from our curated StyleGAN, StyleGAN2, StyleGAN2-ADA, and StyleGAN3 datasets. As such, we ended up with a balanced real-fake dataset with 4,000 samples in each class.

\subsection{Salience Entropy ($S_{entropy}$)}


$S_{entropy}$ scores how much model uncertainty is stored in a variable (in our case, in a Grad-CAM salience heatmap). Table \ref{tab:results} results show that, on average, human-guided models have more compact salience maps (lower $S_{entropy}$ scores and much tighter standard deviations) when compared to cross-entropy models. In particular, the DenseNet model trained with the human-guided loss function recorded the smallest $S_{entropy}$ our of all tested models, what indicates that its salience maps tend to have less noise and be more concentrated when compared to other other models. This also resonates with the elevated consistency of Grad-CAM salience maps for human-guided models, as discussed later in the context of salience stability in Sec. \ref{Salience Stability}. 


\subsection{Salience-Assessed Reaction to Reaction To Noise ($S_{noise}$)}

The values of $S_{noise}$ calculated for two architectures (DenseNet and ResNet) and two types of training (cross-entropy and human-aided) are summarized in Tab. \ref{tab:results}. Values closer to 1.0 denote higher similarity between salience maps for clear and degraded (noisy) samples. This measure allows to conclude interesting properties of the models used in this example. The human-guided training regularizes both models (DenseNet and ResNet) in a way that their salience is equally robust against  added noise (``salt and pepper'' in this case). This can be concluded from similar values of $S_{noise}$ in the last two rows of the $S_{noise}$ section in Tab. \ref{tab:results}. However, training with regular cross-entropy loss produces a model with either large differences between salience maps for clean and noisy samples, or a model that is not affected by the noise.


\begin{figure*}[!htb]
  \begin{subfigure}[t]{.5\textwidth}
    \centering
    \includegraphics[width=\linewidth]{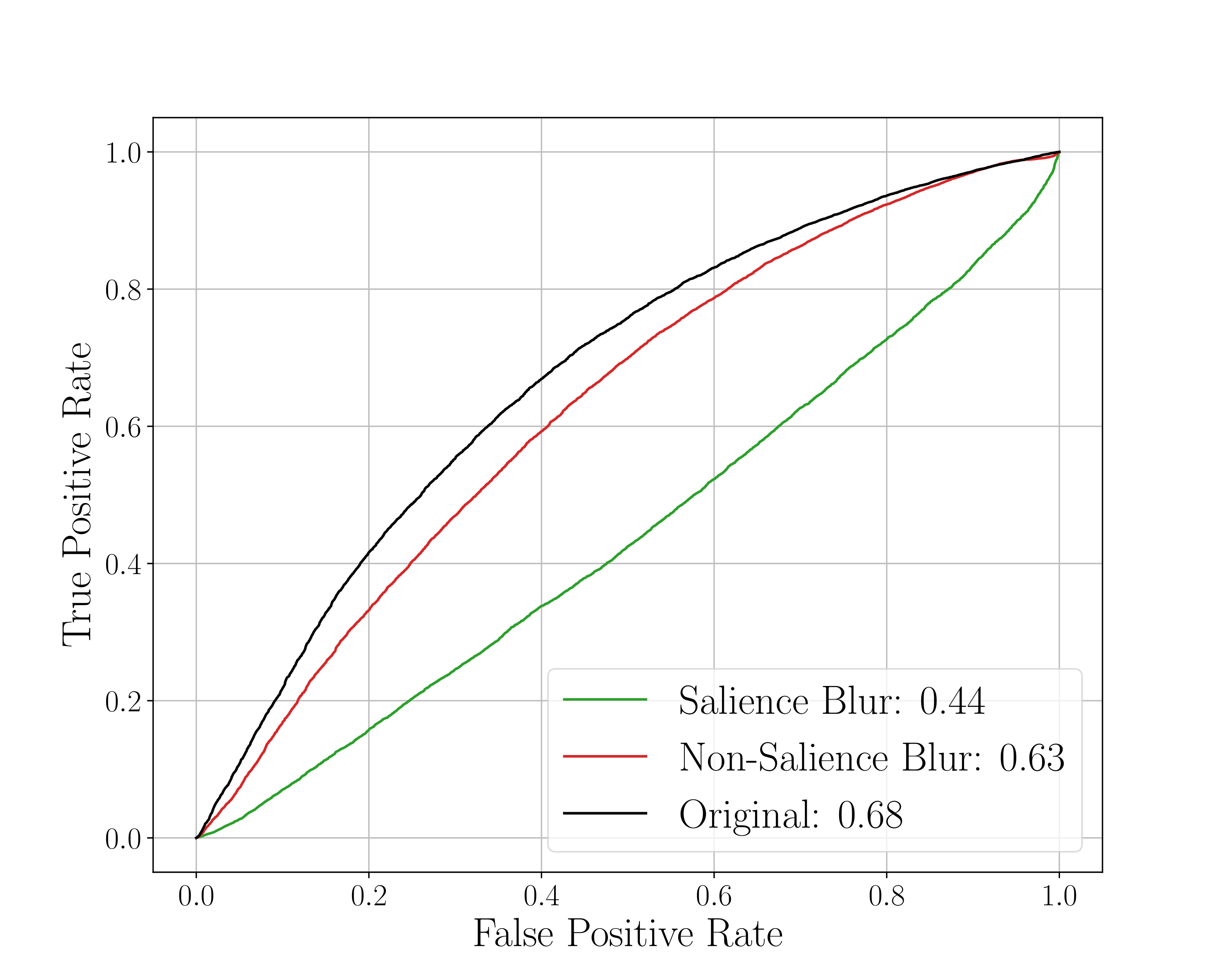}
    \caption{DenseNet / Cross-entropy loss}
  \end{subfigure}
  \hfill
  \begin{subfigure}[t]{.5\textwidth}
    \centering
    \includegraphics[width=\linewidth]{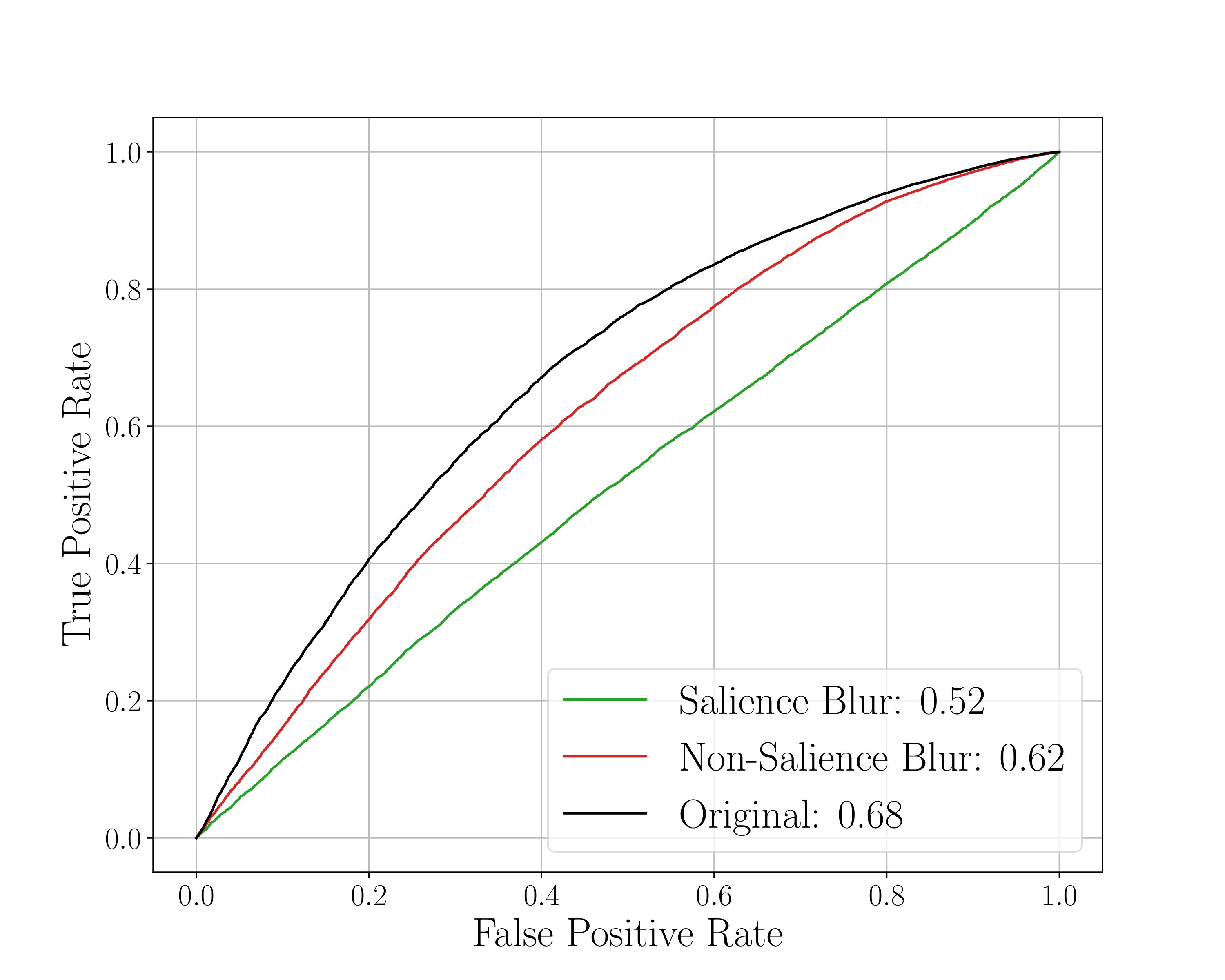}
    \caption{DenseNet / Human-guided loss}
  \end{subfigure}

  \medskip

  \begin{subfigure}[t]{.5\textwidth}
    \centering
    \includegraphics[width=\linewidth]{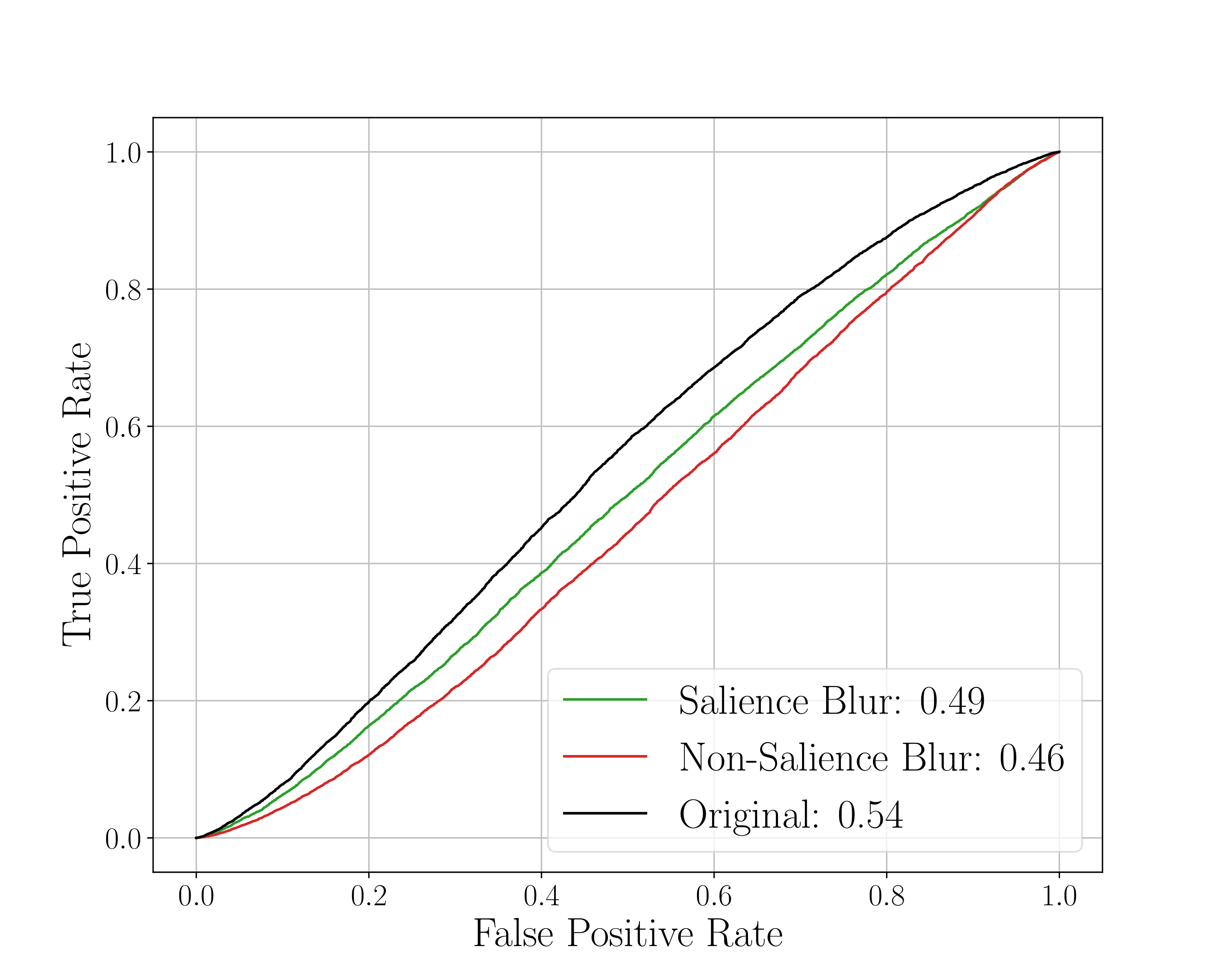}
    \caption{ResNet / Cross-entropy loss}
  \end{subfigure}
  \hfill
  \begin{subfigure}[t]{.5\textwidth}
    \centering
    \includegraphics[width=\linewidth]{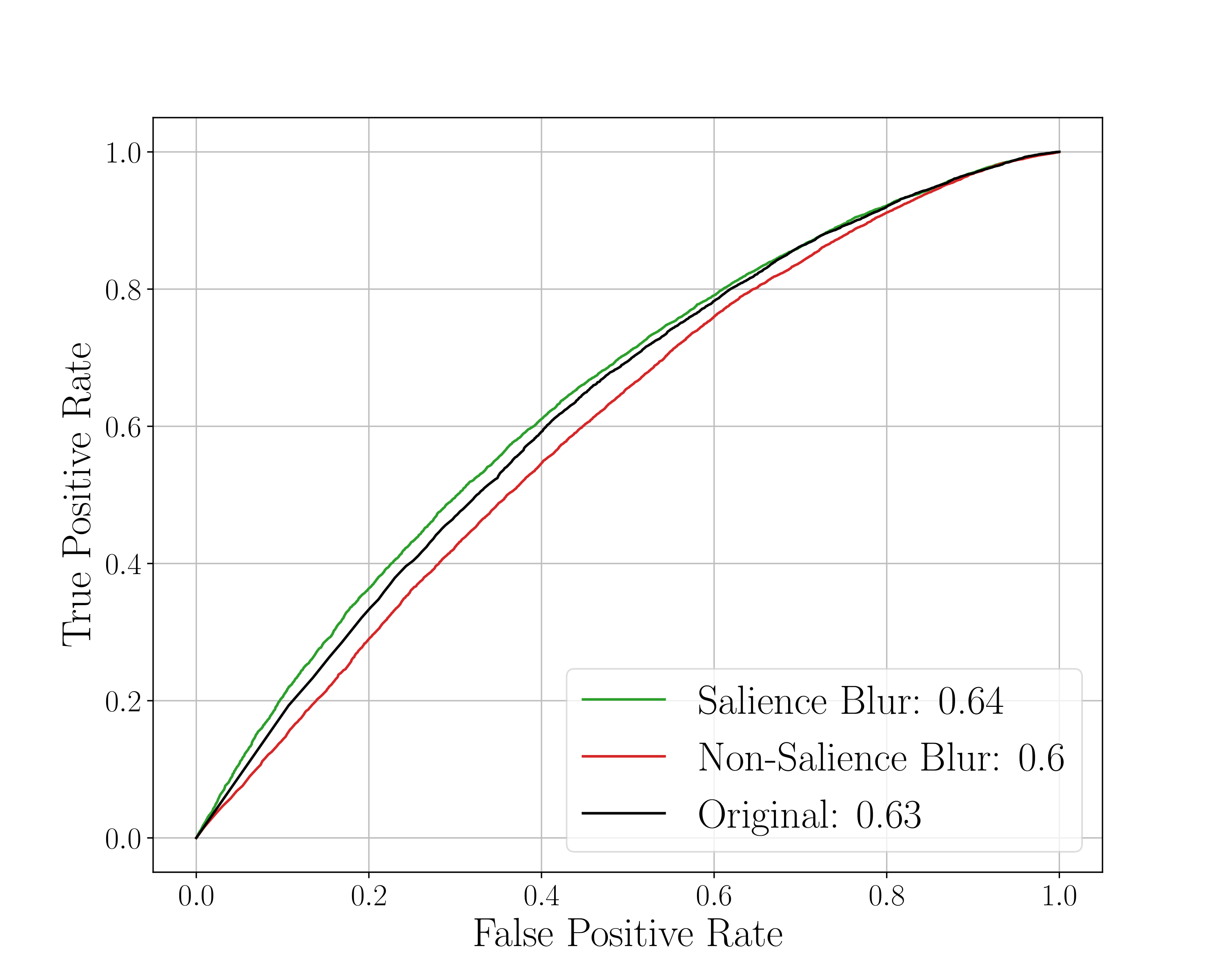}
    \caption{ResNet / Human-guided loss}
  \end{subfigure}
  \caption{Areas Under ROC curves (AUROC) for all four training-architecture configurations, obtained for original samples, after salient region (object-related) removal, and after non-salient region (background) removal.}
  \label{fig:AUROC}
\end{figure*}

\subsection{Salience Resilience to Geometrical Transformations ($S_{resilience}$)}



Table \ref{tab:resilience} shows values of $S_{resilience}$ for the four model architecture-training strategy configurations. To simplify this table, we present values averaged within each group of geometrical transformations (shifts, flips and rotations). It is interesting to see that one architecture (DenseNet) obtains similar $S_{resilience}$ in corresponding categories of transformations independently of the training strategy (regular cross-entropy or human saliency-based). This is not the case for another model (ResNet), which better follows the placement of salient features when the model is trained with human-guided manner. This demonstrates how $S_{resilience}$ can be utilized. Assuming that all four model architecture-training configurations demonstrate similar performance, $S_{resilience}$ suggests that DenseNet should better preserve the model's salience under selected geometrical transformations.





\begin{figure*}[htb]
  \begin{subfigure}[b]{1\textwidth}
      \begin{subfigure}[b]{0.45\textwidth}
          \centering
            \includegraphics[width=1\columnwidth]{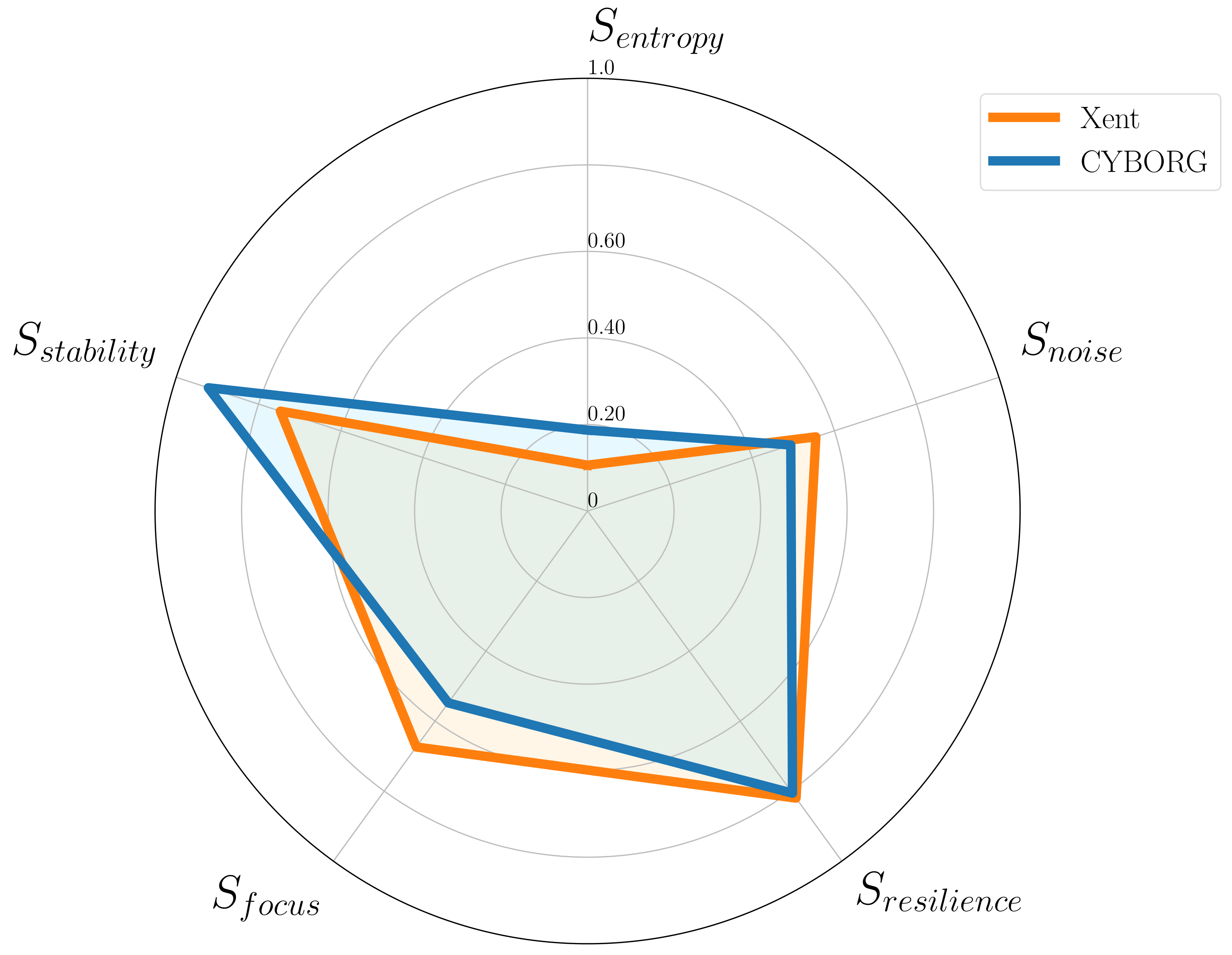}
          \caption{DenseNet}
      \end{subfigure}
      \hfill
      \begin{subfigure}[b]{0.45\textwidth}
          \centering
          \includegraphics[width=1\columnwidth]{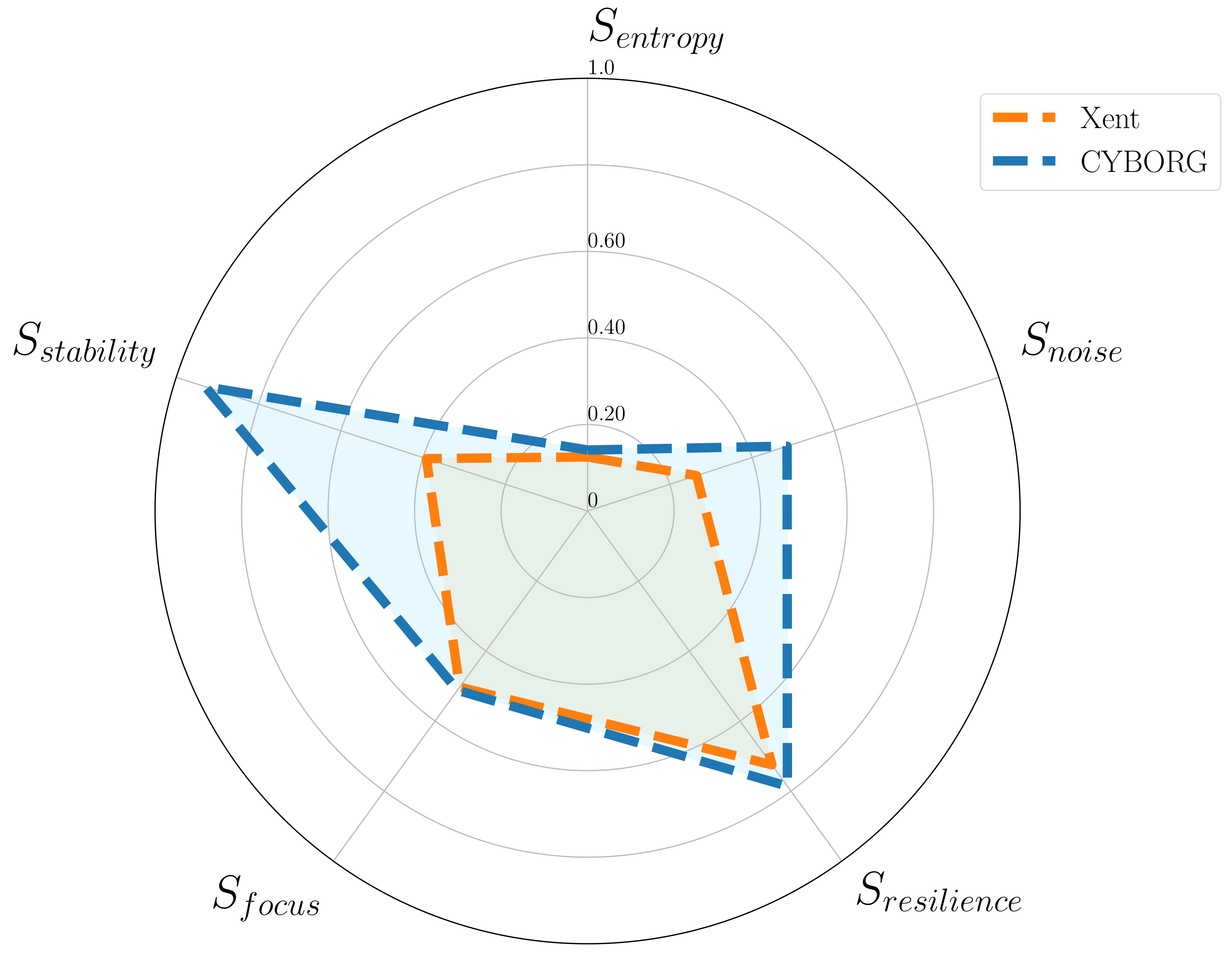}
          \caption{ResNet}
      \end{subfigure}
  \end{subfigure}\vskip3mm
  \caption{Graphical representation of the models (DenseNet and ResNet) and training types (Xent and CYBORG) using all of the proposed explainability measures in this paper. All measures are already normalized between 0 and 1, but in the case of $S_{focus}$, Salience removal was reversed (1 - $S_{focus}$) and averaged with Non-salience removal to reflect the appropriate polarity. $S_{entropy}$ was also reversed (1 - $S_{entropy}$) since lower values of $S_{entropy}$ reflect more focused model salience. Both DenseNet models perform similarly for $S_{noise}$ and $S_{resilience}$, but different for $S_{entropy}$, $S_{stability}$, and $S_{focus}$. On the contrary, both ResNet models were competitive $S_{entropy}$, $S_{resilience}$ and $S_{focus}$, but had large differences for $S_{stability}$ and $S_{noise}$. Our proposed measures indicate the various strengths and weaknesses of each model, allowing for more meaningful and human-explainable information to be derived than purely performance metrics.}
  \label{fig:pentagon_vis}
  \null\vskip-5mm
\end{figure*}

\subsection{Salience-Based Image Degradation ($S_{focus}$)}

Salience-based input sample degradation can be first examined through the lens of the performance. Fig. \ref{fig:AUROC} shows ROC curves (with corresponding Areas Under the Curve value) for the four combinations of model and training strategy used in this paper. In general, we should expect a large decrease in accuracy when the salient regions are removed, and slightly lower performance when non-salient regions are removed, depending on the role the background plays in a given task. 

Interestingly, such behavior is seen for one model (DenseNet, Fig. \ref{fig:AUROC} (a) and (b)), but not for the other (ResNet, Fig. \ref{fig:AUROC} (c) and (d)). While DenseNet suffers significantly from blurring the salient regions (AUROC drops from 0.68 to 0.44 and 0.52, for cross-entropy and human-aided losses, respectively), ResNet is less impacted by the removal of salient information, and the performance of model trained with human-aided loss is on par for original and modified input samples. 

This is where $S_{focus}$ measure can offer explanation related to model's salient regions observed before and after image modifications. Tab. \ref{tab:results} quantifies how the model's salience changed for degraded samples. From that table we see that similarity between the original salience map and the new one obtained after removal of salient parts of the images, is lower on average for DenseNet that for ResNet. It means that the former model (DenseNet) focused on significantly different parts of the images after degradation, what apparently for a problem at hand was not a good strategy. Conversely, the latter model (ResNet) was more robust against salience removal, with the maximum $S_{focus}$ obtained for human-aided training. When, instead of model-wise analyses, we look at the training strategies, we see -- according to the expectations -- that values of $S_{focus}$ are higher in general (independently of salient or non-salient information removal) for models trained with the use of human perception.


\subsection{Salience Stability Across Training Runs ($S_{stability}$)}

Finally, to evaluate the usefulness of the $S_{stability}$ measure, for each architecture-training variant, ten models were independently trained. Salience maps were then generated for all images, both real and fake, rendering a collection of ten salience maps for each image (one for each trained model), as illustrated earlier in Fig. \ref{fig:stability_illustration}. 


The quantitative results are summarized in Table. \ref{tab:results}. From the higher $S_{stability}$ values it is clear that the human-guided training ends up with models vastly outperforming the cross-entropy-trained models across real and synthetic data sets in terms of salience stability. For ten independently trained models that share the same architecture and training data, models guided by human perception during training have more similar salience maps across all training runs (for the same test samples). In contrast, cross-entropy-trained models tend to have great variation in model salience. This is further echoed by the smaller standard deviations of $S_{stability}$ for human-guided models compared to models trained classically. This aligns well with one of the goals of human-aided training, which was to make the model more agnostic to non-salient features accidentally correlated with class labels found during the stochastic training process. And shows that this measure can be also used to assess the effectiveness of the training process.


\subsection{Comparing Models Graphically}
Fig. \ref{fig:pentagon_vis} shows graphically the evaluation of the models (DenseNet and ResNet) and training types (Xent and CYBORG) using all of the proposed explainability measures in this paper. All measures are normalized between 0 and 1, and in the case of $S_{focus}$, Salience removal was reversed (1 - $S_{focus}$) and averaged with Non-salience removal to reflect the appropriate polarity. $S_{entropy}$ was also reversed (1 - $S_{entropy}$ since lower values of $S_{entropy}$ reflect more focused model salience. These graphs can be used to explore model strengths and weaknesses across the five explainability measures developed in this paper. Though the DenseNet models have the same performance (0.68), they differ in explainability measures $S_{stability}$ and $S_{focus}$. Conversely, the ResNet Xent and CYBORG models have wildly different performance (0.54 and 0.63, respectively), they achieve similar explainability for measures $S_{resilience}$ and $S_{focus}$.



\section{Discussion}

In this paper, we broaden the utility of model salience by introducing a variety of methods to assess their role in model performance and their human explainability. The set of five introduced measures, all solely based on model's salience, includes assessment of: (a) model's salience entropy (to understand a general model's spatial selectivity of features), (b) reaction to noise (to understand the model's focus as a function of input degradation level), (c) resilience to geometrical transformations (to understand whether the spatial location of salient features follows the transformations applied to the input), (d) how the removal of important and background features impacts the new salience generated by the model, and finally (e) what is the impact of different random seeds on the shape of the salience. A core contribution of this work is that the proposed measures offer single-number quantitative assessments, which can be generated for arbitrarily-large test datasets, and thus complement usually low-at-scale qualitative assessments utilizing class activation maps.

To demonstrate how the proposed measures work in practice, we performed our metrics on two popular deep CNN architectures (DenseNet and ResNet), trained with two different strategies (using classical cross-entropy loss, and incorporating human perceptual intelligence into training). Since these four architecture-training combinations produce models behaving differently in terms of their salience, that was a convenient way to assess the usefulness of the proposed measures. It is, however, important to note that the proposed salience-based assessment ideas are not specific to these variants. Also, the types of input deformations applied in this work (\eg random noise in $S_{noise}$, or particular geometrical transformations in $S_{resilience}$, or the number of training runs in $S_{stability}$) are only illustrative examples, suitable for synthetic face detection problem. Other problems may benefit from other deformations (\eg Gaussian or salt and pepper noise can be replaced with synthetically-generated atmospheric turbulence patterns in long-range face recognition problem) and successfully apply the proposed explainability measures. To facilitate the application of this work, the source codes are offered along with this paper.

{
\small
\bibliographystyle{ieee}
\bibliography{main}

\begin{thebibliography}{10}\itemsep=-1pt

\bibitem{angelov2021explainable}
P.~P. Angelov, E.~A. Soares, R.~Jiang, N.~I. Arnold, and P.~M. Atkinson.
\newblock Explainable artificial intelligence: an analytical review.
\newblock {\em Wiley Interdisciplinary Reviews: Data Mining and Knowledge
  Discovery}, 11(5):e1424, 2021.

\bibitem{arnold2019factsheets}
M.~Arnold, R.~K. Bellamy, M.~Hind, S.~Houde, S.~Mehta, A.~Mojsilovi{\'c},
  R.~Nair, K.~N. Ramamurthy, A.~Olteanu, D.~Piorkowski, et~al.
\newblock Factsheets: Increasing trust in ai services through supplier's
  declarations of conformity.
\newblock {\em IBM Journal of Research and Development}, 63(4/5):6--1, 2019.

\bibitem{SaliencyMapsWithoutSacrificingAccuracy}
V.~Aswal, G.~Kao, S.~Y. Kim, and K.~Morrison.
\newblock Towards generating human-centered saliency maps without significantly
  sacrificing accuracy.
\newblock {\em NeuroVision2022: What can computer vision learn from visual
  neuroscience? Workshop at CVPR 2022}, 2022.

\bibitem{bengfort_yellowbrick_2018}
B.~Bengfort, R.~Bilbro, N.~Danielsen, L.~Gray, K.~{McIntyre}, P.~Roman, Z.~Poh,
  et~al.
\newblock Yellowbrick, 2018.

\bibitem{boyd2022human}
A.~Boyd, K.~W. Bowyer, and A.~Czajka.
\newblock Human-aided saliency maps improve generalization of deep learning.
\newblock In {\em Proceedings of the IEEE/CVF Winter Conference on Applications
  of Computer Vision}, pages 2735--2744, 2022.

\bibitem{Boyd_2023_WACV_CYBORG}
A.~Boyd, P.~Tinsley, K.~W. Bowyer, and A.~Czajka.
\newblock Cyborg: Blending human saliency into the loss improves deep
  learning-based synthetic face detection.
\newblock In {\em Proceedings of the IEEE/CVF Winter Conference on Applications
  of Computer Vision (WACV)}, pages 6108--6117, January 2023.

\bibitem{chattopadhay2018grad}
A.~Chattopadhay, A.~Sarkar, P.~Howlader, and V.~N. Balasubramanian.
\newblock Grad-cam++: Generalized gradient-based visual explanations for deep
  convolutional networks.
\newblock In {\em 2018 IEEE winter conference on applications of computer
  vision (WACV)}, pages 839--847. IEEE, 2018.

\bibitem{dovsilovic2018explainable}
F.~K. Do{\v{s}}ilovi{\'c}, M.~Br{\v{c}}i{\'c}, and N.~Hlupi{\'c}.
\newblock Explainable artificial intelligence: A survey.
\newblock In {\em 2018 41st International convention on information and
  communication technology, electronics and microelectronics (MIPRO)}, pages
  0210--0215. IEEE, 2018.

\bibitem{Fong_ICCV_2017}
R.~C. Fong and A.~Vedaldi.
\newblock Interpretable explanations of black boxes by meaningful perturbation.
\newblock In {\em 2017 IEEE International Conference on Computer Vision
  (ICCV)}, pages 3449--3457, 2017.

\bibitem{axiom_cam}
R.~Fu, Q.~Hu, X.~Dong, Y.~Guo, Y.~Gao, and B.~Li.
\newblock Axiom-based grad-cam: Towards accurate visualization and explanation
  of cnns.
\newblock {\em arXiv preprint arXiv:2008.02312}, 2020.

\bibitem{goldstein2015peeking}
A.~Goldstein, A.~Kapelner, J.~Bleich, and E.~Pitkin.
\newblock Peeking inside the black box: Visualizing statistical learning with
  plots of individual conditional expectation.
\newblock {\em journal of Computational and Graphical Statistics},
  24(1):44--65, 2015.

\bibitem{ResNet}
K.~He, X.~Zhang, S.~Ren, and J.~Sun.
\newblock Deep residual learning for image recognition.
\newblock In {\em 2016 IEEE Conference on Computer Vision and Pattern
  Recognition (CVPR)}, pages 770--778, 2016.

\bibitem{huang2017densely}
G.~Huang, Z.~Liu, L.~Van Der~Maaten, and K.~Q. Weinberger.
\newblock Densely connected convolutional networks.
\newblock In {\em Proceedings of the IEEE conference on computer vision and
  pattern recognition}, pages 4700--4708, 2017.

\bibitem{layer_cam}
P.-T. Jiang, C.-B. Zhang, Q.~Hou, M.-M. Cheng, and Y.~Wei.
\newblock Layercam: Exploring hierarchical class activation maps for
  localization.
\newblock {\em IEEE Transactions on Image Processing}, 30:5875--5888, 2021.

\bibitem{Karras2020ada}
T.~Karras, M.~Aittala, J.~Hellsten, S.~Laine, J.~Lehtinen, and T.~Aila.
\newblock Training generative adversarial networks with limited data.
\newblock In {\em Proc. NeurIPS}, 2020.

\bibitem{StyleGAN3}
T.~Karras, M.~Aittala, S.~Laine, E.~H{\"a}rk{\"o}nen, J.~Hellsten, J.~Lehtinen,
  and T.~Aila.
\newblock Alias-free generative adversarial networks.
\newblock {\em Advances in Neural Information Processing Systems}, 34:852--863,
  2021.

\bibitem{karras2019style}
T.~Karras, S.~Laine, and T.~Aila.
\newblock A style-based generator architecture for generative adversarial
  networks.
\newblock In {\em Proceedings of the IEEE/CVF conference on computer vision and
  pattern recognition}, pages 4401--4410, 2019.

\bibitem{StyleGAN2}
T.~Karras, S.~Laine, M.~Aittala, J.~Hellsten, J.~Lehtinen, and T.~Aila.
\newblock Analyzing and improving the image quality of stylegan.
\newblock In {\em Proceedings of the IEEE/CVF conference on computer vision and
  pattern recognition}, pages 8110--8119, 2020.

\bibitem{DeepGaze}
A.~Linardos, M.~K{\"u}mmerer, O.~Press, and M.~Bethge.
\newblock Deepgaze iie: Calibrated prediction in and out-of-domain for
  state-of-the-art saliency modeling.
\newblock In {\em Proceedings of the IEEE/CVF International Conference on
  Computer Vision}, pages 12919--12928, 2021.

\bibitem{lundberg2017unified}
S.~M. Lundberg and S.-I. Lee.
\newblock A unified approach to interpreting model predictions.
\newblock {\em Advances in neural information processing systems}, 30, 2017.

\bibitem{molnar2020interpretable}
C.~Molnar.
\newblock {\em Interpretable machine learning}.
\newblock Lulu. com, 2020.

\bibitem{IS_cam}
R.~Naidu, A.~Ghosh, Y.~Maurya, S.~S. Kundu, et~al.
\newblock Is-cam: Integrated score-cam for axiomatic-based explanations.
\newblock {\em arXiv preprint arXiv:2010.03023}, 2020.

\bibitem{noyes2017super}
E.~Noyes, P.~J. Phillips, and A.~O’Toole.
\newblock What is a super-recogniser?
\newblock In {\em Face processing: Systems, disorders and cultural
  differences}, pages 173--201. Nova Science Publishers Inc, 2017.

\bibitem{smooth_cam}
D.~Omeiza, S.~Speakman, C.~Cintas, and K.~Weldermariam.
\newblock Smooth grad-cam++: An enhanced inference level visualization
  technique for deep convolutional neural network models.
\newblock {\em arXiv preprint arXiv:1908.01224}, 2019.

\bibitem{ramaswamy2020ablation}
H.~G. Ramaswamy et~al.
\newblock Ablation-cam: Visual explanations for deep convolutional network via
  gradient-free localization.
\newblock In {\em Proceedings of the IEEE/CVF Winter Conference on Applications
  of Computer Vision}, pages 983--991, 2020.

\bibitem{lime}
M.~T. Ribeiro, S.~Singh, and C.~Guestrin.
\newblock "why should {I} trust you?": Explaining the predictions of any
  classifier.
\newblock In {\em Proceedings of the 22nd {ACM} {SIGKDD} International
  Conference on Knowledge Discovery and Data Mining, San Francisco, CA, USA,
  August 13-17, 2016}, pages 1135--1144, 2016.

\bibitem{richards2020methodology}
J.~Richards, D.~Piorkowski, M.~Hind, S.~Houde, and A.~Mojsilovi{\'c}.
\newblock A methodology for creating ai factsheets.
\newblock {\em arXiv preprint arXiv:2006.13796}, 2020.

\bibitem{cam_entropy2022}
A.~Schöttl.
\newblock Improving the interpretability of gradcams in deep classification
  networks.
\newblock {\em Procedia Computer Science}, 200:620--628, 2022.
\newblock 3rd International Conference on Industry 4.0 and Smart Manufacturing.

\bibitem{selvaraju2017grad}
R.~R. Selvaraju, M.~Cogswell, A.~Das, R.~Vedantam, D.~Parikh, and D.~Batra.
\newblock Grad-cam: Visual explanations from deep networks via gradient-based
  localization.
\newblock In {\em Proceedings of the IEEE international conference on computer
  vision}, pages 618--626, 2017.

\bibitem{tjoa2020survey}
E.~Tjoa and C.~Guan.
\newblock A survey on explainable artificial intelligence (xai): Toward medical
  xai.
\newblock {\em IEEE transactions on neural networks and learning systems},
  32(11):4793--4813, 2020.

\bibitem{UK_HoL_2018}
{UK House of Lords, Select Committee on Artificial Intelligence}.
\newblock {AI in the UK: ready, willing and able?}, 2018.
\newblock Report of Session 2017–19.

\bibitem{SS_cam}
H.~Wang, R.~Naidu, J.~Michael, and S.~S. Kundu.
\newblock Ss-cam: Smoothed score-cam for sharper visual feature localization.
\newblock {\em arXiv preprint arXiv:2006.14255}, 2020.

\bibitem{score_cam}
H.~Wang, Z.~Wang, M.~Du, F.~Yang, Z.~Zhang, S.~Ding, P.~Mardziel, and X.~Hu.
\newblock Score-cam: Score-weighted visual explanations for convolutional
  neural networks.
\newblock In {\em Proceedings of the IEEE/CVF conference on computer vision and
  pattern recognition workshops}, pages 24--25, 2020.

\bibitem{Wang_TIP_2004}
Z.~Wang, A.~Bovik, H.~Sheikh, and E.~Simoncelli.
\newblock Image quality assessment: from error visibility to structural
  similarity.
\newblock {\em IEEE Transactions on Image Processing}, 13(4):600--612, 2004.

\bibitem{wexler2019if}
J.~Wexler, M.~Pushkarna, T.~Bolukbasi, M.~Wattenberg, F.~Vi{\'e}gas, and
  J.~Wilson.
\newblock The what-if tool: Interactive probing of machine learning models.
\newblock {\em IEEE transactions on visualization and computer graphics},
  26(1):56--65, 2019.

\bibitem{activation_based_cam}
B.~Zhou, A.~Khosla, A.~Lapedriza, A.~Oliva, and A.~Torralba.
\newblock Learning deep features for discriminative localization.
\newblock In {\em Proceedings of the IEEE conference on computer vision and
  pattern recognition}, pages 2921--2929, 2016.

\end{thebibliography}
}

\end{document}